\pdfoutput=1

\documentclass[11pt]{article}

\usepackage[]{EMNLP2022}

\usepackage{times}
\usepackage{latexsym}

\usepackage[T1]{fontenc}

\usepackage{booktabs}
\usepackage{tabularx}
\usepackage[textsize=scriptsize]{todonotes}
\usepackage{multirow}
\usepackage{url}
\usepackage{amsmath}
\usepackage{mathtools}
\usepackage{amssymb}
\usepackage{amsthm}
\usepackage{tablefootnote}
\usepackage{caption}
\usepackage{subcaption}
\usepackage{bbm}
\usepackage{comment}
\usepackage{xspace}
\usepackage{algorithm}
\usepackage{proof}
\usepackage{stmaryrd}
\usepackage{bm}
\usepackage{pifont}

\usepackage[utf8]{inputenc}

\usepackage{microtype}

\usepackage{inconsolata}
\usepackage[textsize=scriptsize]{todonotes}

\newcommand{\mnli}{MNLI}
\newcommand{\qqp}{QQP}
\newcommand{\msgs}{MSGS}
\newcommand{\hans}{HANS}
\newcommand{\paws}{PAWS}

\newcommand{\verbset}{\textsc{VERB}}
\newcommand{\morph}{\textsc{MORPH}}
\newcommand{\adj}{\textsc{ADJECT}}

\newcommand{\hanssub}{\textsc{HANS-SUB}}
\newcommand{\hanscon}{\textsc{HANS-CON}}
\newcommand{\accuracy}{\texttt{ACC}}
\newcommand{\confidence}{\texttt{CONF}}

\newcommand\defeq{\mathrel{\overset{\makebox[0pt]{\mbox{\normalfont\tiny\sffamily def}}}{=}}}

\title{Assessing Out-of-Domain Language Model Performance from Few Examples}

\author{Prasann Singhal$^*$, Jarad Forristal$^*$, Xi Ye, and Greg Durrett \\
  Department of Computer Science \\
The University of Texas at Austin \\
    \texttt{ \{prasanns, jarad, xiye, gdurrett\}@cs.utexas.edu}}

\usepackage{enumitem}

\usepackage{xargs}                      
\newcommandx{\unsure}[2][1=]{\todo[linecolor=red,backgroundcolor=red!25,bordercolor=red,#1]{#2}}
\newcommandx{\change}[2][1=]{\todo[linecolor=blue,backgroundcolor=blue!25,bordercolor=blue,#1]{#2}}

\begin{document}
\maketitle
\def\thefootnote{*}\footnotetext{Equal contribution}\def\thefootnote{\arabic{footnote}}

\begin{abstract}

While pretrained language models have exhibited impressive generalization capabilities, they still behave unpredictably under certain domain shifts. In particular, a model may learn a reasoning process on in-domain training data that does not hold for out-of-domain test data. We address the task of predicting out-of-domain (OOD) performance in a few-shot fashion: given a few target-domain examples and a set of models with similar training performance, can we understand how these models will perform on OOD test data? We benchmark the performance on this task when looking at model accuracy on the few-shot examples, then investigate how to incorporate analysis of the models' behavior using feature attributions to better tackle this problem. Specifically, we explore a set of ``factors'' designed to reveal model agreement with certain pathological heuristics that may indicate worse generalization capabilities. On textual entailment, paraphrase recognition, and a synthetic classification task, we show that attribution-based factors can help rank relative model OOD performance. However, accuracy on a few-shot test set is a surprisingly strong baseline, particularly when the system designer does not have in-depth prior knowledge about the domain shift.
\end{abstract}

\section{Introduction}

The question of whether models have learned the right behavior on a training set is crucial for generalization. Deep models have a propensity to learn shallow reasoning shortcuts \cite{Geirhos2020} like single-word correlations \cite{gardner-etal-2021-competency} or predictions based on partial inputs \cite{poliak-etal-2018-hypothesis}, particularly for problems like natural language inference \cite{gururangan-etal-2018-annotation,hans} and question answering 
\cite{jia-liang-2017-adversarial,chen-durrett-2019-understanding}. 
Unless we use evaluation sets tailored to these spurious signals, accurately understanding if a model is learning them remains a hard problem ~\cite{NLPshortcuts, kim2021, generalsurvey}. 

\begin{figure}[t]
\centering
\includegraphics[width=\linewidth, trim=10 255 280 20,clip]{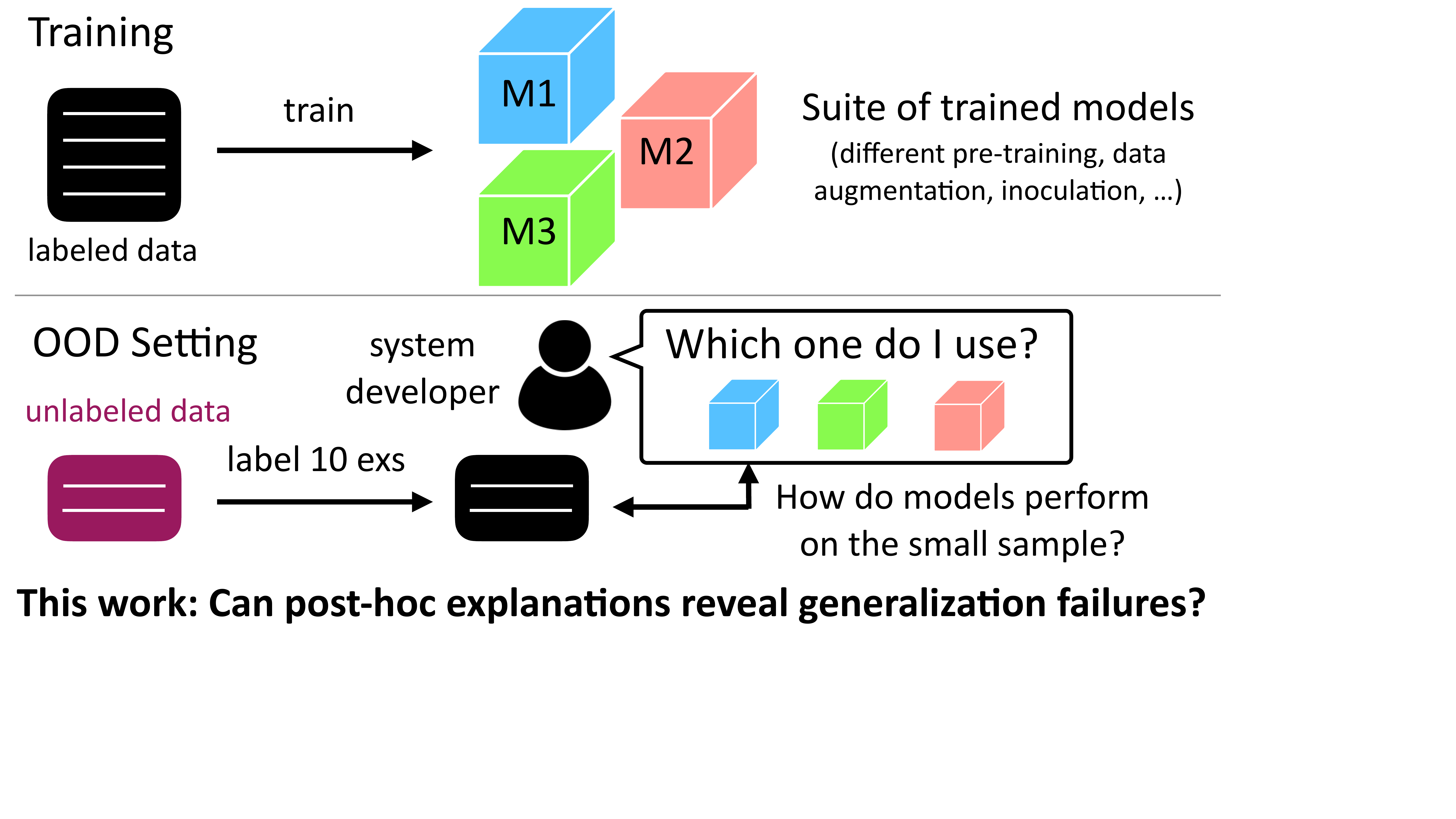}
\caption{Our setting: a system developer is trying to evaluate a collection of trained models on a small amount of hand-labeled data to assess which one may work best in this new domain. Can baselines / attributions help?}
\vspace{-3mm}
\label{fig:framework}
\end{figure}

This paper addresses the problem of \emph{predicting} whether a model will work well in a target domain \emph{given only a few examples from that domain.} This setting is realistic: a system designer can typically hand-label a few examples to serve as a test set. Computing accuracy on this small set and using that as a proxy for full-test set performance is a simple baseline for our task, but has high variance, which may cause us to incorrectly rank two models that achieve somewhat similar performance. We hypothesize that we can do better if we can interpret the model's behavior beyond accuracy. With the rise of techniques to analyze post-hoc feature importance in machine-learned models~\cite{lundberg2017unified, lime, tokig}, we have seen not just better interpretation of models, but improvements such as constraining them to avoid using certain features \cite{RossEtAl2017rightforright} like those associated with biases \cite{liu-avci-2019-incorporating,kennedy-etal-2020-contextualizing}, or trying to more generally teach the right reasoning process for a problem \cite{Yao2021RefiningLM,tang21-rightforright,pruthi2022}. If post-hoc interpretation can strengthen a models' ability to generalize, can they also help us understand it?

Figure~\ref{fig:framework} illustrates the role this understanding can play. We have three trained models and are trying to rank them for suitability on a new domain. The small labeled dataset is a useful (albeit noisy) indicator of success. However, by checking model attributions on our few OOD samples, we can more deeply understand model behavior and analyze if they use certain pathological heuristics. Unlike past work \cite{adebayo2022post}, we seek to automate this process as much as possible, provided the unwanted behaviors are characterizable by describable heuristics. We use scalar factors, which are simple functions of model attributions, to estimate proximity to these heuristics, similar to characterizing behavior in past work \cite{ye-etal-2021-connecting}. We then evaluate whether these factors allow us to correctly rank the models' performance on OOD data.

Both on synthetic \cite{msgs}, and real datasets \cite{hans,paws}, we find that, between models with similar architectures but different training processes, both our accuracy baseline and attribution-based factors are good at distinguishing relative model performance on OOD data. However, on models with different base architectures, we discovering interesting patterns, where factors can very strongly distinguish between different types of models, but cannot always map these differences to correct predictions of OOD performance. In practice, we find probe set accuracy to be a quick and reliable tool for understanding OOD performance, whereas factors are capable of more fine-grained distinctions in certain situations. 


\paragraph{Our Contributions:} (1) We benchmark, in several settings, methods for predicting and understanding relative OOD performance with few-shot OOD samples. (2) We establish a ranking-based evaluation framework for systems in our problem setting. (3) We analyze patterns in how accuracy on a few-shot set and factors derived from token attributions distinguish models.

\begin{figure}[t!]
\centering
\includegraphics[width=\linewidth, trim=25 210 680 25,clip]{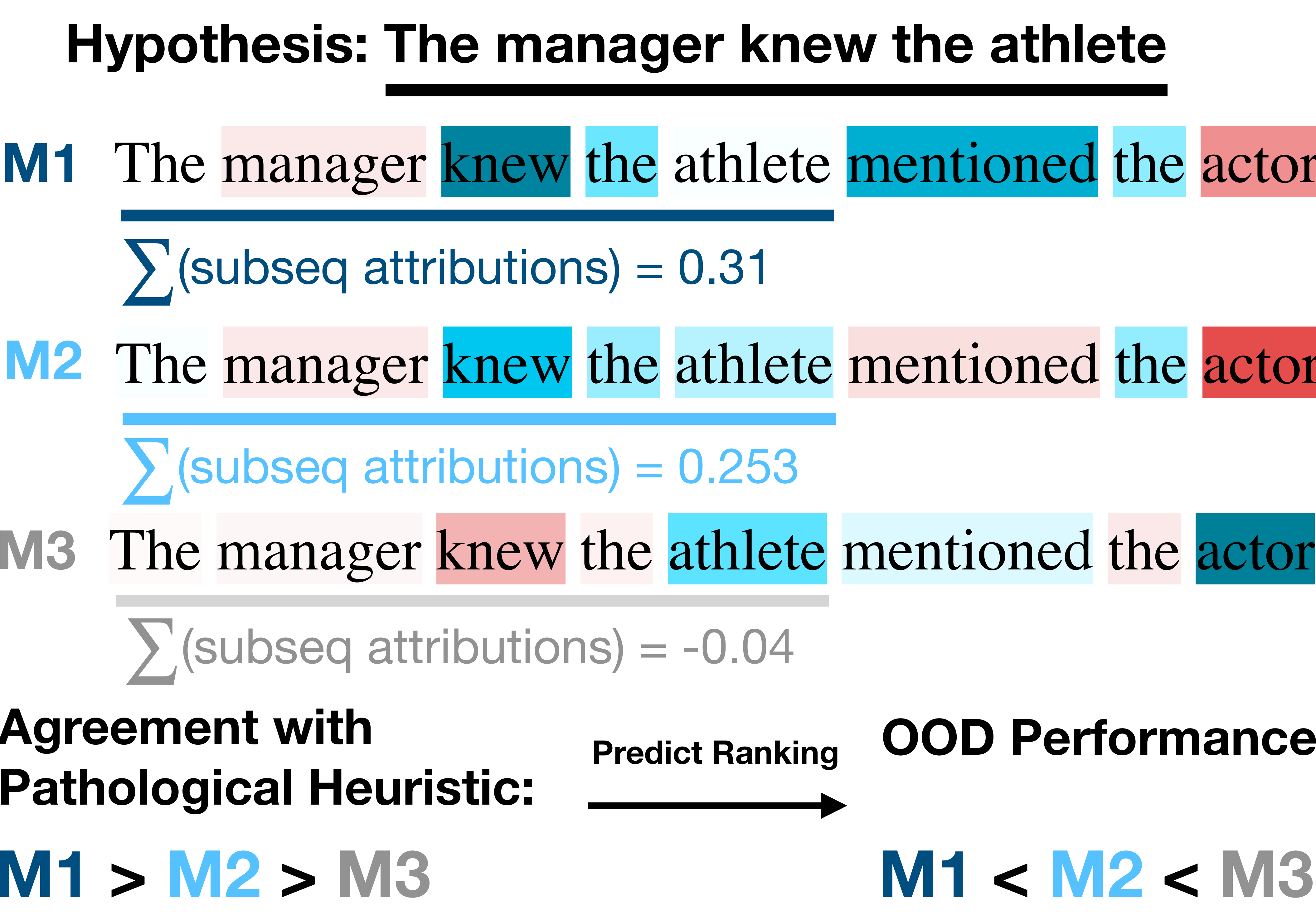}

\caption{Explanations generated on the same sample for \hans{} subsequence data models M1, M2, M3 (have ascending OOD performance). The factor (shaded underlines) from knowledge of the OOD allows us to in this example predict the model ranking.}
\vspace{-3mm}
\label{fig:hansexample}
\end{figure}

\section{Motivating Example}
\label{sec:motivating}

To expand on Figure~\ref{fig:framework}, Figure~\ref{fig:hansexample} shows an in-depth motivating example of our process. We show three feature attributions from three different models on an example from the HANS dataset~\cite{hans}. These models have (unknown) varied OOD performance but similar performance on the in-domain MNLI~\cite{mnli} data. Our task is then to correctly rank these models' performance on the HANS dataset in a few-shot manner.

We can consider ranking these models via simple metrics like accuracy on the small few-shot dataset, where higher-scoring models are higher-ranked. However, such estimates can be high variance on small datasets. In Figure~\ref{fig:hansexample}, only M3 predicts non-entailment correctly, and we cannot distinguish the OOD performance of M1 and M2 without additional information.

Thus, we turn to explanations to gain more insight into the models' underlying behavior. With faithful attributions, we should be able to determine if the model is following simple inaccurate rules called \emph{heuristics}~\cite{hans}. Figure~\ref{fig:hansexample} shows the heuristic where a model predicts that the sentence $A$ entails $B$ if $B$ is a subsequence of $A$. Crucially, we can use model attributions to assess model use of this heuristic :we can sum the attribution mass the model places on subsequence tokens. We use the term \emph{factors} to refer to such functions over model attributions. 

The use of factors potentially allows for the automation of detection of spurious signals or shortcut learning~\cite{Geirhos2020}. While prior work has shown that spurious correlations are hard for a human user to detect from explanations~\cite{adebayo2022post}, well-designed factors could automatically analyze model behavior across a number of tasks and detect such failures.


\section{Attributions to Predict Performance}\label{sec:explanations_ood}

In this section, we formalize the ideas presented thus far. Token-level attribution methods (a subset of post-hoc explanations) are methods which, given an input sequence of tokens $\mathbf{x} \defeq x_1, x_2, ... , x_n$ and a model prediction $\hat{y} \defeq M(\mathbf{x})$ for some task, assign an explanation $\phi(\mathbf{x},\hat{y}) \defeq a_1,\ldots,a_n$ where $a_i$ corresponds to an attribution or importance score for a corresponding $x_i$ towards the final prediction. For cases where the model, prediction, and inputs are unambiguous, we abbreviate this simply $\phi_i  \equiv \phi(\mathbf{x}) \defeq \phi(\mathbf{x}, M_i(\mathbf{x}))$.

We assume that the model is trained on an in-domain training dataset $D_T$ and will be evaluated on some unknown OOD set $D_O$. Given two models $M_0$ and $M_1$, with a small amount of data $D_{(O, t)} \subset D_O$ ($t = 10$ examples or fewer in our settings), our task is to predict which model will generalize better. We break the process into 2 steps (see Figure~\ref{fig:hansexample}):

\paragraph{1. Hypothesize a heuristic.} First we must identify an underlying heuristic $H$ that reflects pathological model behavior in the OOD dataset. For example, the subsequence heuristic in Figure~\ref{fig:hansexample} corresponds to a heuristic which always predicts entailed if the hypothesis is contained within the premise. Let $h(M_i)$ abstractly reflect how closely the $i$th model's behavior aligns with $H$. Let $s(M_i)$ be the true OOD performance of model $M_i$. If we then assume that $h(M_i)$ faithfully models some pathological heuristic $H$, we should have that $h(M_0) > h(M_1) > \ldots > h(M_m)$ implies $s(M_0) < s(M_1) < \ldots < s(M_m)$ . In other words, the more a model $M_i$ agrees with a pathological heuristic $H$, the worse it performs.

\paragraph{2. Measure alignment.} We now want to predict the ranking of $s(M_i)$; however, with few labeled examples there may be high variance in directly evaluating these metrics. We instead use factors $f(\mathbf{x}, \phi_i)$ which map tokens and their attributions for model $M_i$ to scalar scores that should correlate with the heuristic $H$. Factors can be designed to align with known pathological heuristics, where higher scores indicate strong model agreement with the associated heuristic. We then estimate the ranking of $s(M_i)$ using the relative ranking of the corresponding $h(M_i)$ approximated through factors. 

Concretely, to measure the alignment, we first compute for each input $\mathbf{x}_j \in D_{(O, t)}$ the prediction $M_i(\mathbf{x}_j$) and the explanation $\phi(\mathbf{x}_j)$ for that prediction. These $\phi(\mathbf{x}_j)$ are used to compute the score $f(\mathbf{x}_j,\phi(\mathbf{x}_j))$ for model $M$. We take the overall score of the model to be $F(i) = \frac{1}{t} \sum_{j=1}^t f(\mathbf{x}_j,\phi(\mathbf{x}_k, M_i(\mathbf{x}_k)))$, the mean over the $t$ examples in $D_{(O, t)}$. We then directly rank models on the basis of the $F(i)$ values: the higher the average factor value (the more it follows the heuristic), the lower the relative ranking: $F(0) > F(1) \implies s(M_0) < s(M_1)$. Therefore we can sort the models by these values and arrive at a predicted ranking. We later also consider factors which to not intuitively map to specific heuristics.





\paragraph{Baselines} We also consider three principle explanation-agnostic baselines. A natural baseline given $D_{(O, t)}$ is to simply use the accuracy (\accuracy{}) on this dataset: $\frac{1}{n} \sum_{i=1}^n \mathbbm{1}[y_i = M(\mathbf{\mathbf{x}_i})$], however this may be noisy on only a few examples, and frequently leads to ties.\footnote{Most of the datasets we consider are constructed specifically to mislead models following the heuristic, so this baseline directly measures agreement with a heuristic $h$.}

We can also assess model confidence (\confidence{}), which looks at the softmax probability of the predicted label, as well as looking at \texttt{CONF-GT} which only looks at the softmax probability of the ground-truth label. 

\section{Experimental Setup}

\subsection{Models Compared}
\label{sec:models_compared}

In this work, we compare various models across different axes yielding different $D_O$ performance. The first approach we use is \textbf{inoculation} \cite{liu-etal-2019-inoculation}, which involves fine-tuning models on small amounts or batches of $D_O$ data alongside in-domain data to increase model performance on OOD data. The second approach we use is varying the model architecture and pre-training (e.g., using a stronger pre-trained Transformer model). 

In Section~\ref{sec:msgs}, we use inoculation to create 5 RoBERTa-base~\cite{roberta} models of varying $D_O$ performance for each of the three \msgs{} sets. In Section~\ref{sec:realistic} where we consider the \hans{} and \paws{} datasets, we inoculate a variety of models. For \hans{}, we inoculate 5 RoBERTa-large models. We additionally examine DeBERTa-v3-base~\cite{deberta, debertav3} and ELECTRA-base~\cite{electra} models fine-tuned on in-domain \textsc{MNLI} data. For \paws{}, we inoculate 4 RoBERTa-base models on the in-domain $D_T$ set. We also inoculate ELECTRA-base and DEBERTA-base models. We include complete details for these models in Appendix~\ref{appendix:inoc_details}. The generated models represent a realistic problem scenario: a practitioner may have many different models with similar $D_T$ performance, but different $D_O$ performance. We specifically crafted suites of models which have both near pairs (models with similar $D_O$ performance) and far pairs.

\subsection{Attribution Methods}


We experiment using several token-level attributions methods: \textbf{LIME} \cite{lime} computes attribution scores using the weights of a linear model approximating model behavior near a datapoint. \textbf{SHAP} \cite{lundberg2017unified} is similar to LIME, but uses a procedure using Shapley values. Finally, \textbf{Integrated Gradients (\textsc{tokig})} \cite{tokig} compute $\phi_i$ by performing a line integral over the gradients with respect to token embeddings on a path from a baseline token to the ground truth token; commonly, this baseline token is chosen to be \texttt{<MASK>}. While intuitively sensible, \citet{harbecke2021explaining} has voiced concerns regarding the use of \textsc{TOKIG} in NLP.

\subsection{Evaluation Setup}
\label{sec:fewshotmetric}

Because model ranking using a small $D_{(O, t)}$ may be unstable, we conduct all experiments over a number of different sampled $D_{(O, t)}$ sets. We first sample $M$ examples from each set (in the range of 200-600), then generate explanations for all models on each example. We then take 400-500 bootstrap samples of size $n$ (we report results for $n = 10$, as experimental results were similar for sizes 5 and 20), simulating many few-shot evaluations. For each bootstrap sample, we analyse ${m \choose 2}$ model pairs. Details can be found in Appendix~\ref{appendix:bootstrapdets}.

We define a ``success'' as a technique correctly ranking a model pair, when measured by $D_O$ performance (on the full set); otherwise is a ``failure''. We define \emph{pairwise accuracy} as the accuracy for a method ranking a particular model pair across all bootstrap samples. We define \emph{few-shot accuracy} (or just \emph{accuracy}) as the average of the pairwise accuracies over the ${m \choose 2}$ model pairs. By reporting ranking accuracy across a diverse set of models, we ensure a comprehensive evaluation.  

\section{MSGS: A Proof of Concept}\label{sec:msgs}

\begin{figure}[t!]
\centering
\includegraphics[width=\linewidth,trim=0mm 95mm 300mm 10mm]{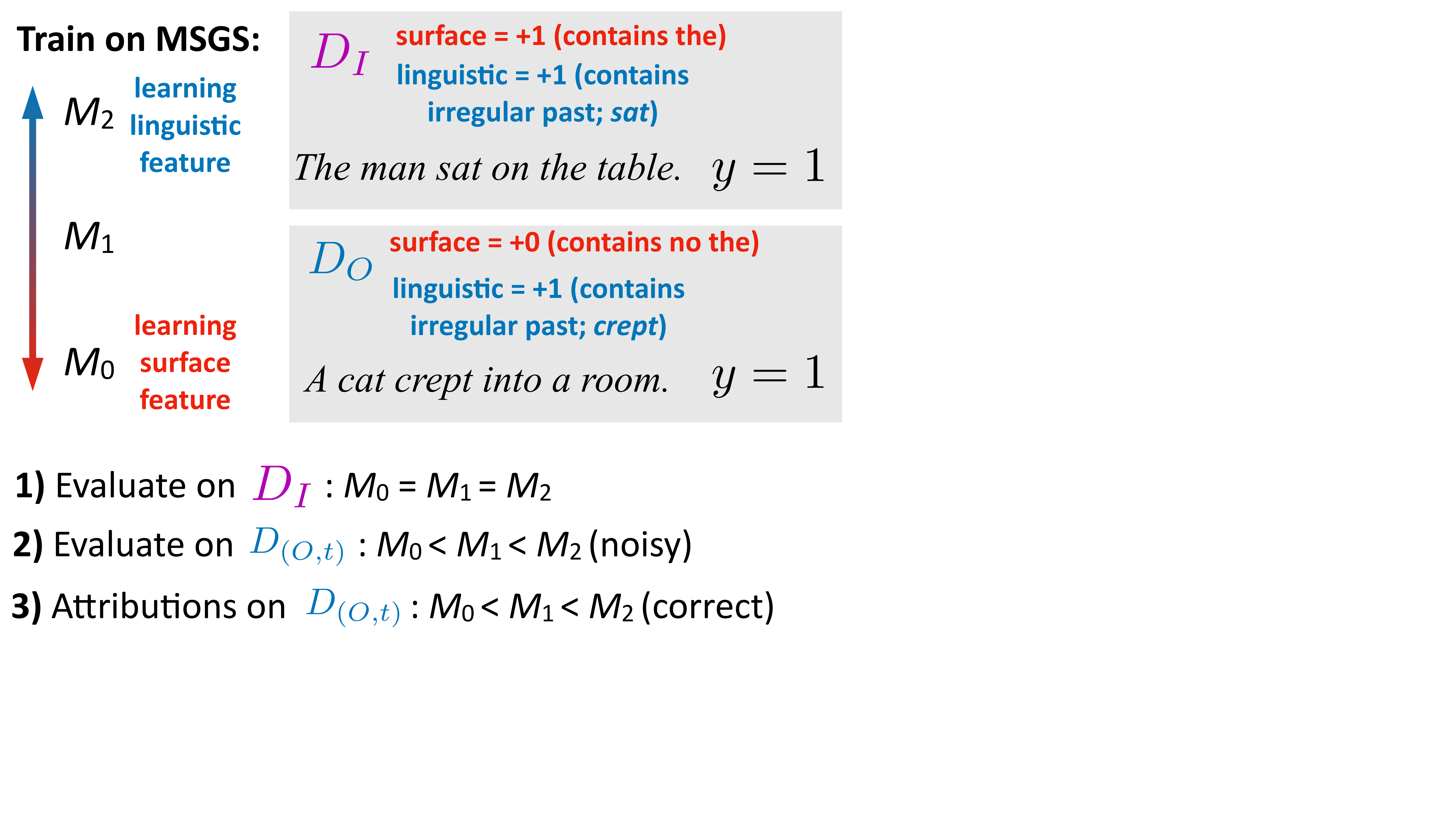}
\caption{Example from the MSGS train and OOD test sets. The training data conflates a surface and linguistic generalization as described in \citet{msgs}, resulting in models that learn a range of behaviors. Direct evaluation OOD on small data can tell us this, but explanations can also differentiate which of the two patterns is learned and how strongly they are learned.}

\label{fig:msgs-fig}
\end{figure}

We first show experiments on the Mixed Signals Generalization Set (\msgs) dataset presented in~\citet{msgs} as a proof of concept for our methodology. \msgs{} is a synthetic classification dataset. The training (in-domain) set is composed of sentences where both some linguistic feature (e.g., the presence of an adjective) and a spurious surface feature (e.g., the word ``\emph{the}'' being in the sentence) are always associated with a positive label $y=1$. This data is ambiguous, which means the model could rely on either the linguistic or surface feature completely yet still get 100\% accuracy on in-domain data. \citet{msgs} then create sets of OOD data where the linguistic feature becomes associated with the $y=1$ positive label, and the surface feature with a $y=0$ label. The resulting test accuracy reflects model reliance on one feature or the other. \citet{msgs} use this to investigate what generalizations are learned at which stages of model pre-training; 
we investigate whether information from small probe sets can help assess model reliance on the surface feature.

We consider three of their linguistic features: \morph{} (presence of an irregular past verb like ``\emph{drew}''), \adj{} (prescence of an adjective), and \verbset{} (if the main verb is an -ing verb), each paired with the surface feature of ``\emph{the}'' being in the sentence.

We design factors which look at attributions on the tokens corresponding to these linguistic features, including the tokens surrounding these features as well to account for feature dependence on surrounding words. Our factor $f(\mathbf{x},\phi) = -\sum_{i=(m-2)}^{m+2} \phi(x_i)$, where $m$ is the index of the feature-critical word for that dataset (e.g., ``\emph{slept}'' for \texttt{IRREG}) and $\phi(x_i)$ is the attribution at an index. This factor corresponds closely to the heuristic that the dataset was designed for, or alternately, we can see this factor as inversely proportional to what \emph{other} information the model is using (that is, information outside of this window). We name the factors \texttt{IRREG}, \texttt{VERG}, and \texttt{ADJ} for the \morph{}, \verbset{}, \adj{} sets respectively.

Note that this approach assumes that a system designer has prior knowledge of the relevant linguistic and surface feature. This is a generous assumption, and for this dataset is almost sufficient to formulate the rule used to construct it, hence why we call this a proof of concept. We will show more realistic conditions in Section~\ref{sec:realistic}.

\paragraph{Models} To create a suite of models with varying $D_O$ performance, we inoculate following the steps outlined in Section~\ref{sec:models_compared}. We evaluate our factors via accuracy as described in Section~\ref{sec:fewshotmetric}. More details about the inoculation is present in Section~\ref{appendix:inoc_details} of the appendix.


\begin{table}[!t]
    \centering
    \small
    \begin{tabular}{cl|ccc}
    
        \toprule
        \multirow{5}{*}{}
        Feature & Method & \multicolumn{3}{c}{Accuracy}\\
         
         \midrule
         
         \multirow{5}{*}{\morph{}}
         & {\texttt{ACC}} & \multicolumn{3}{c}{90.9} \\
         & {\texttt{CONF}} & \multicolumn{3}{c}{50.9} \\
         & {\texttt{CONF-GT}} & \multicolumn{3}{c}{90.1} \\
         \cmidrule{3-5}
          & & {\sc tokig} &  {\sc shap} & {\sc lime}  \\
         & {\texttt{IRREG}}  & 89.2 & 90.6 & \bf 92.8$\dagger$ \\
         \midrule
         \multirow{5}{*}{\verbset{}}
         & {\texttt{ACC}} & \multicolumn{3}{c}{94.5}\\
         & {\texttt{CONF}} & \multicolumn{3}{c}{58.0} \\
         & {\texttt{CONF-GT}} & \multicolumn{3}{c}{93.3} \\
         \cmidrule{3-5}
         & & {\sc tokig} &  {\sc shap} & {\sc lime}  \\
         & {\texttt{VERB}} &  92.1 &  94.0 & \bf 94.9 \\
         \midrule
         \multirow{5}{*}{\adj{}}
         & {\texttt{ACC}} & \multicolumn{3}{c}{89.9}\\
         & {\texttt{CONF}} & \multicolumn{3}{c}{50.5}\\
         & {\texttt{CONF-GT}} & \multicolumn{3}{c}{91.3} \\
         \cmidrule{3-5}
         & & {\sc tokig} &  {\sc shap} & {\sc lime}  \\
         &{\texttt{ADJ}} & 87.4 & 92.1 & \bf 93.5$\dagger$ \\
         \bottomrule
    \end{tabular}
    \caption{Few-shot ranking accuracy metric results on $D_{(O, t)}$ for \msgs{}. \texttt{IRREG}, \texttt{VERB}, and \texttt{ADJ} are detailed in Section~\ref{sec:msgs}. $\dagger$ indicates statistically significant improvement over accuracy (paired bootstrap test: $p<0.05$)}
    \label{tab:msgs_indistr}
\end{table}

\paragraph{Results} Table~\ref{tab:msgs_indistr}  shows the results on this dataset. Our \texttt{ACC} baseline performs well: when models differ greatly in performance (e.g., one gets 50\% and another gets 90\% on the $D_O$), accuracy on the small $D_{(O, t)}$ ranks these correctly even despite the small subset size. The high regularity of the dataset also means that a model's behavior does not vary greatly from example to example, further reducing variance. However, this ranking is nevertheless still not perfect. We see that \texttt{CONF} performs very poorly, by contrast, showing that confidence is not helpful for measuring model behavior.

Overall, we see that methods using explanations are able to beat the \texttt{ACC} baseline, with the exception of \textsc{tokig}. We additionally found trends within the explanation techniques themselves, with \textsc{lime} reliably performing the best, and \textsc{tokig} being the worst. But generally, all techniques can offer relevant information, and in the best case, the attributions can tell us more reliably what a model is learning than evaluation on a small set of $D_{(O, t)}$ data can. In Section~\ref{sec:realistic}, we investigate if these results generalize to real-world datasets.

\section{Realistic OOD Settings}
\label{sec:realistic}

We now consider two datasets corresponding to realistic OOD settings treated in past work. 

First, \textbf{HANS}~\cite{hans} targets spurious heuristics within MNLI~\cite{mnli}, such as the hypothesis being a subsequence of the premise, with balanced test sets that can be used to detect model reliance on these heuristics. Models following these heuristics always predict \emph{entailed} for the hypotheses, and will perform at random chance accuracy on the dataset. We use MNLI as our in-domain training set in this setting.

Second, \textbf{PAWS}~\cite{paws} is a paraphrase identification task. PAWS-QQP is an OOD dataset for Quora Question Pairs (QQP)~\cite{qqp} that is composed of pairs with swapped content words/phrases (e.g., \emph{I ran from the Grand Canyon to California} to \emph{I ran from California to the Grand Canyon}). A paraphrase model that relies heavily on lexical overlap will not be sensitive to these changes, and will always predict the label of $y=1$ to indicate paraphrase. We use QQP as our in-domain training set in this setting.

Details regarding models used in this section are presented in Section~\ref{sec:models_compared}. From the test sets of the corresponding datasets, we randomly sample 400 examples from \paws{} and 600 from \hanscon{} and \hanssub{} each for use in bootstrap sampling, as detailed in Section~\ref{sec:fewshotmetric}. Information regarding the datasets considered can be found in Table~\ref{tab:data_settings}.

\subsection{Factors}

\paragraph{General Factors} Both \hans{} and \paws{} involve comparing two sequences $\mathbf{a}, \mathbf{b}$ of tokens, unlike \msgs{} which is classification over a single sequence. We can define our input $\mathbf{x} = a_1, a_2, ... a_n, b_1, b_2, ..., b_m$ as composed of these two sequences $\mathbf{a}$ and $\mathbf{b}$ with respective attributions $\mathbf{\phi}_a, \mathbf{\phi}_b$. We evaluate a number of factors that generally target sensitivity to both sequences and their differences, which represent a broad class of potential heuristics.

\begin{enumerate}[align=left, itemsep=-.5em]
    \item[{\texttt{MAX-DIFF}}:] The difference between maximum attribution in $\mathbf{a}$ and $\mathbf{b}$, i.e. $\max(\phi_a) - \max(\phi_b)$.
    \item[{\texttt{SUM-DIFF}}:] the difference of the sum of attributions, i.e. $\sum_{i=1}^n \phi_{a,i} - \sum_{i=1}^m \phi_{b,i}$.
    \item[{\texttt{INDEX-DIFF}}:] The difference of attributions between shared words in $\mathbf{a}$ and $\mathbf{b}$.
    \item[{\texttt{FIRST-TOK}}:] The attribution at the the first \texttt{<SEP>} token.
\end{enumerate}
We explicitly note that this is the exhaustive set of factors we experimented with, not a cherry-picked set, in order to provide a comprehensive view of what does and doesn't work. We crafted these by manually examining attribution patterns on various datasets rather than trying a large number and keeping the best ones.




\paragraph{HANS Factors} We look at the ``subsequence'' heuristic discussed in Section~\ref{sec:motivating} and the constituent heuristic, which assumes that the premise entails all complete subtrees in its parse-tree. For the subsequence OOD set (\hanssub{}) we note that the \texttt{INDEX-DIFF} factor, which specifically examines tokens in the shared subsequence, captures the setting's pathological heuristic. 

On the constituent OOD set (\hanscon{}) we evaluate a factor that examines the attribution on the control words of the premise. For example, for the premise ``\emph{Unless the doctors ran, the lawyers encouraged the scientists}'' and the hypothesis ``\emph{The doctors ran}'', we would consider the attributions on the word ``\emph{Unless}''.

\begin{table}[t]
    \centering

    \small
    \begin{tabular}{lccc}
            \toprule
            
             & \multicolumn{3}{c}{$D_{(O, t)}$}\\
             \cmidrule{2-4}
             
            \textbf{Ranking Method} &  \multicolumn{1}{c}{\paws{}} & \multicolumn{1}{c}{\hanssub} & \multicolumn{1}{c}{\hanscon} \\
          \midrule
         \bf Baselines    \\
         \accuracy{} &  88.7 & 90.6 & 81.6   \\
         \confidence{} & \phantom{0}9.2 & 40.4 & 52.8   \\
         {\texttt{CONF-GT} } & 34.9 & 20.2 & 38.9   \\
         {\texttt{RANDOM} } & 50.7 & 51.4 & 49.6   \\
         \midrule
         \bf Factors   \\
         {\texttt{CONST}} & $-$  & $-$ & \bf 87.1  \\
         {\texttt{SWAP-MAX-DIFF}}  & 76.2 & $-$ & $-$   \\
         {\texttt{SWAP-AVG}} & \bf 91.4 & $-$ & $-$  \\
         {\texttt{INDEX-DIFF}} & 60.5  & \bf 91.3 & 68.6   \\
         \midrule
         {\texttt{MAX-DIFF}} & 65.9 & 69.2 & 50.6  \\
         {\texttt{SUM-DIFF}} &  56.6 & 60.0 & 75.2   \\
         {\texttt{FIRST-TOK}} & 74.3 & 50.4  & 55.6   \\
         \bottomrule
    \end{tabular}
    \caption{Few-shot heuristic ranking performance on OOD samples for \hans{}/\mnli{} and \qqp{}/\paws{}, specifically when comparing \textbf{inoculated models} (SHAP explanations). We divide rows by baselines, dataset-specific factors, and general factors. }
    \label{tab:hanspawsfewshotshap}
\end{table}

\paragraph{PAWS Factors}  We further investigate two intuitive heuristics that are based on the construction of the OOD set. \texttt{SWAP-AVG} uses the average attribution across all swapped tokens and \texttt{SWAP-MAX-DIFF} subtracts the highest magnitude attribution of swapped tokens in the first sentence and the highest magnitude attribution of swapped tokens in the second sentence. For example, for the pair (``\emph{What factors cause a good person to become bad ?}'', ``\emph{What factors cause a bad person to become good ?}''), \texttt{SWAP-AVG} would consider the attributions on ``\emph{good}'' and ``\emph{bad}''. \texttt{SWAP-MAX-DIFF} is analogous. 


\subsection{Inoculated Results}

We first evaluate models that differ primarily through inoculation, as described in Section~\ref{sec:models_compared}. Results are shown using SHAP in Table~\ref{tab:hanspawsfewshotshap} which we selected through experiments in this setting as being the best performing. The conclusions here differ somewhat from those on \msgs{}. We note that the \accuracy\ baseline remains strong, while \confidence\ is near random. We find that \emph{certain} attribution factors are able to outperform the \texttt{ACC} baseline, with \texttt{SWAP-AVG} the best on \paws{} (91.4\%), \texttt{INDEX-DIFF} the best on \hanssub{} (91.3\%), and \texttt{CONST} on \hanscon{} (87.1\%).

This shows that even in these settings more realistic than \msgs{}, \textbf{the right choice of factor reveals meaningful information about model generalization}. Moreover, the heuristics that work well are those hand-designed for these datasets, confirming our hypothesis that measuring association with a heuristic via a factor may reveal something about performance. 



We qualify these results by noting that in a true few-shot setting, there is some uncertainty regarding whether a chosen factor is truly the best one. As a coarse option, we find \texttt{ACC} to be reliable. However, these high-performing factors would still be useful in conjunction with accuracy, or if we had previously validated a factor as ranking models well and we wanted to apply it to rank new models in this domain; the factors will generalize to new models even if they do not generalize to new datasets necessarily.

\subsection{Architectural Change Results}

We further examine our approach when ranking the performance of different pre-trained models (RoBERTa, ELECTRA, and DeBERTa). 

Table~\ref{tab:hanspawsshapother} shows that a heuristic {\texttt{GUESS}} based on the expectation across choosing a \emph{best} model and then randomly guessing consistently with that, gives a strong baseline of 72\%. Factors also seem to do well in this setting, with all of the general heuristics outperforming the very low \accuracy{} baseline. 

This suggests that in few-shot factors are able to capture distributional information that baselines can't. However, to qualify this, given that each set only compares between 3 pairs of models, it's easier for factors to happen upon strong accuracy patterns by chance. 


Thus, in Table~\ref{tab:pawspairwiseshap}, we analyze this further by showing resuts on some individual model pairs.
(R1, R2 are RoBERTa; E is Electra; D is DeBERTa) \textsc{R1-E} and \textsc{D-R2}, have different architectures, but similar OOD accuracy (see Table~\ref{tab:innoc_details} in the Appendix). \textsc{R1-D}, \textsc{E-D} and \textsc{E-R2} are different model types with more distant accuracy. Accuracy values for a single pair on this single dataset therefore only reflect differences across bootstrap samples. What we find in common across these types of pairings is that while some values are close to 50\%, including the {\texttt{ACCURACY}} baseline, each column has several factors achieving very distinct (0\%, 100\%) accuracy values, consistently differentiating these models. As we note in Figure~\ref{fig:extremityplot} (Appendix), this pattern of strong distinctions is quite common when different types of models are compared. We further discuss this in Section~\ref{sec:analysis}. 

\begin{table}[t]
    \centering
    \small
    \begin{tabular}{lc}
            \toprule
             & \multicolumn{1}{c}{$D_{(O, t)}$}\\
            \textbf{Ranking Method} & \multicolumn{1}{c}{\textsc{hans/paws pooled}} \\ 
          \midrule
         \bf Baseline    \\
         {\texttt{ACCURACY}} &  61.0   \\
         {\texttt{GUESS}} &  72.0   \\

         \midrule
         \bf Factors   \\
         {\texttt{SET-DEPENDENT}} & 75.5 \\
         \midrule
        
         {\texttt{MAX-DIFF}} & 72.6  \\
         {\texttt{INDEX-DIFF}} & 83.0 \\
         {\texttt{SUM-DIFF}} & 83.1 \\
         {\texttt{FIRST-TOK}} & 69.4 \\
         \bottomrule
    \end{tabular}
    \caption{Few-shot heuristic ranking performance on OOD samples for \hans{}/\mnli{} and \qqp{}/\paws{}, specifically when comparing \textbf{non-inoculated models} (SHAP explanations), where we take the mean of pairwise accuracies for 3 pairs (for 3 models) on each set. }
    \label{tab:hanspawsshapother}
\end{table}

\begin{table}[t]\setlength{\tabcolsep}{4pt}
    \centering
    \small
    \begin{tabular}{lccccc}
            \toprule
            & \multicolumn{5}{c}{Model Pair ($M_1$-$M_2$)} \\
            \cmidrule{2-6}
           \textbf{Ranking Method} &  \multicolumn{1}{c}{\textsc{R1-E}} & \multicolumn{1}{c}{\textsc{R1-D}} & \multicolumn{1}{c}{\textsc{E-R2}}& \multicolumn{1}{c}{\textsc{D-R2}}&
           \multicolumn{1}{c}{\textsc{E-D}}
           \\
          \midrule
         \bf Baseline   \\
         {\texttt{ACCURACY}} &  77.4 & 54.6& 55.2 & 67.2 & 64.4\\
         \midrule 
         \bf Factors   \\
     {\texttt{SWAP-MAX-DIFF}}  & 57.6 & 57.6 & 70.6 & 65.6 & 42.6 \\
         {\texttt{SWAP-AVG}} & 93.4 & 93.4 & 65.0 & 47.6 & 17.0\\
         \midrule
         {\texttt{MAX-DIFF}} & 63.2 & 63.2 & 39.2 & 4.8 & 0.2\\
         {\texttt{INDEX-DIFF}} & 0 & 0 & 99.6 & 99.6 & 15.0\\
         {\texttt{SUM-DIFF}} &  99.8 & 100 & 0 & 0 & 88.0\\
         {\texttt{FIRST-TOK}} & 1.2 & 100  & 99.6 & 0 & 0\\
         \bottomrule
    \end{tabular}
    \caption{SHAP pairwise accuracies, different types of models, PAWS. Model R1 (69.7\%), R2 (82.9\%),  Model E (80.5\%), and Model D (71.8\%)}
    \label{tab:pawspairwiseshap}
\end{table}

\section{Analysis / Discussion}
\label{sec:analysis}




\paragraph{Accuracy is reliable, but factors can provide more fine-grained distinctions.} On \msgs{}, where factors beat strong accuracy baselines, we notice that these pairwise accuracies are consistently high. For example, on the \morph{} setting, for two models with 95\% and 98\% accuracy, our factors \texttt{IRREG} is 100\% accurate, while the accuracy baseline here is only 58\%, as test accuracy on $D_{(O, t)}$ does not discriminate well between two models with such close overall accuracy. 

This holds at the fine-grained pairwise level as well. Figure~\ref{fig:msgspairwise} (also see Figure~\ref{fig:pairwisecomp} in appendix) shows the baseline $D_{(O, t)}$ accuracy against a specific factor's accuracy for each model pair in \msgs{}. Each datapoint in the scatterplot represents a model pair and a point's vertical distance from the red line represents how much better or worse a given factor does compared to the baseline on a specific pair.  We see a regular trend: explanations seem to systematically outperform the baseline across various pairs, with a few significant deviations for low-performing pairs.

These results suggest that explanations can be useful and do add information otherwise missing from accuracy probing alone, especially when the underlying model architecture is held constant. With differing architectures (Figure~\ref{fig:pairwisecomp}), the problem is made more difficult, and selecting the right factor is less obvious; few-shot accuracy may be more reliable in this setting. Note, however, that these successes from any technique are in spite of us only inspecting 10 examples from the target domain.

\begin{figure}[t!]
\centering
\includegraphics[width=0.48\textwidth]{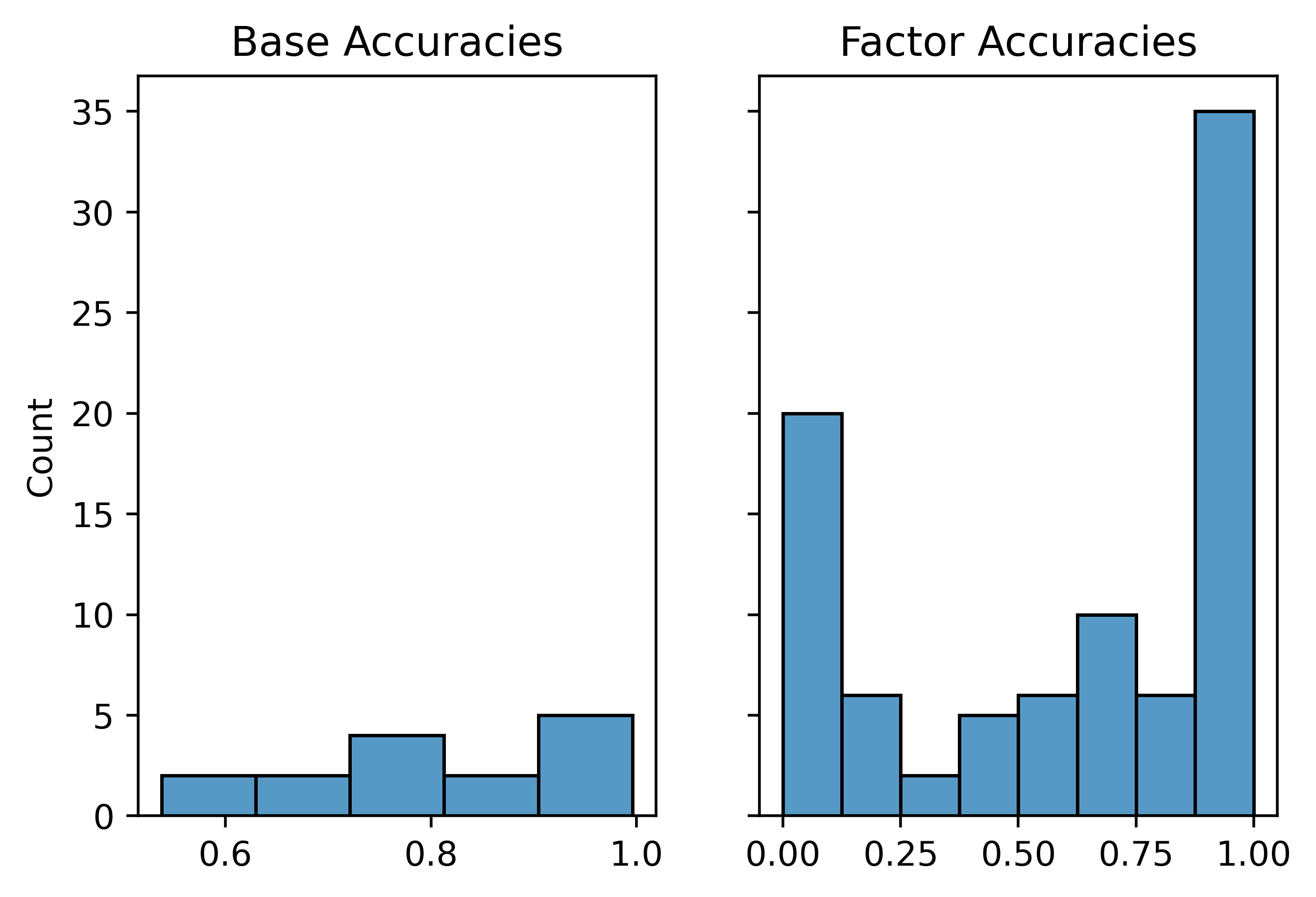}

\caption{Distributions of pairwise accuracies on PAWS SHAP non-inoculated, all model pairs (left for accuracy baseline, right for all factors).} 

\label{fig:extremityplot}
\end{figure}

\begin{figure}[t!]
\centering
\includegraphics[width=0.48\textwidth]{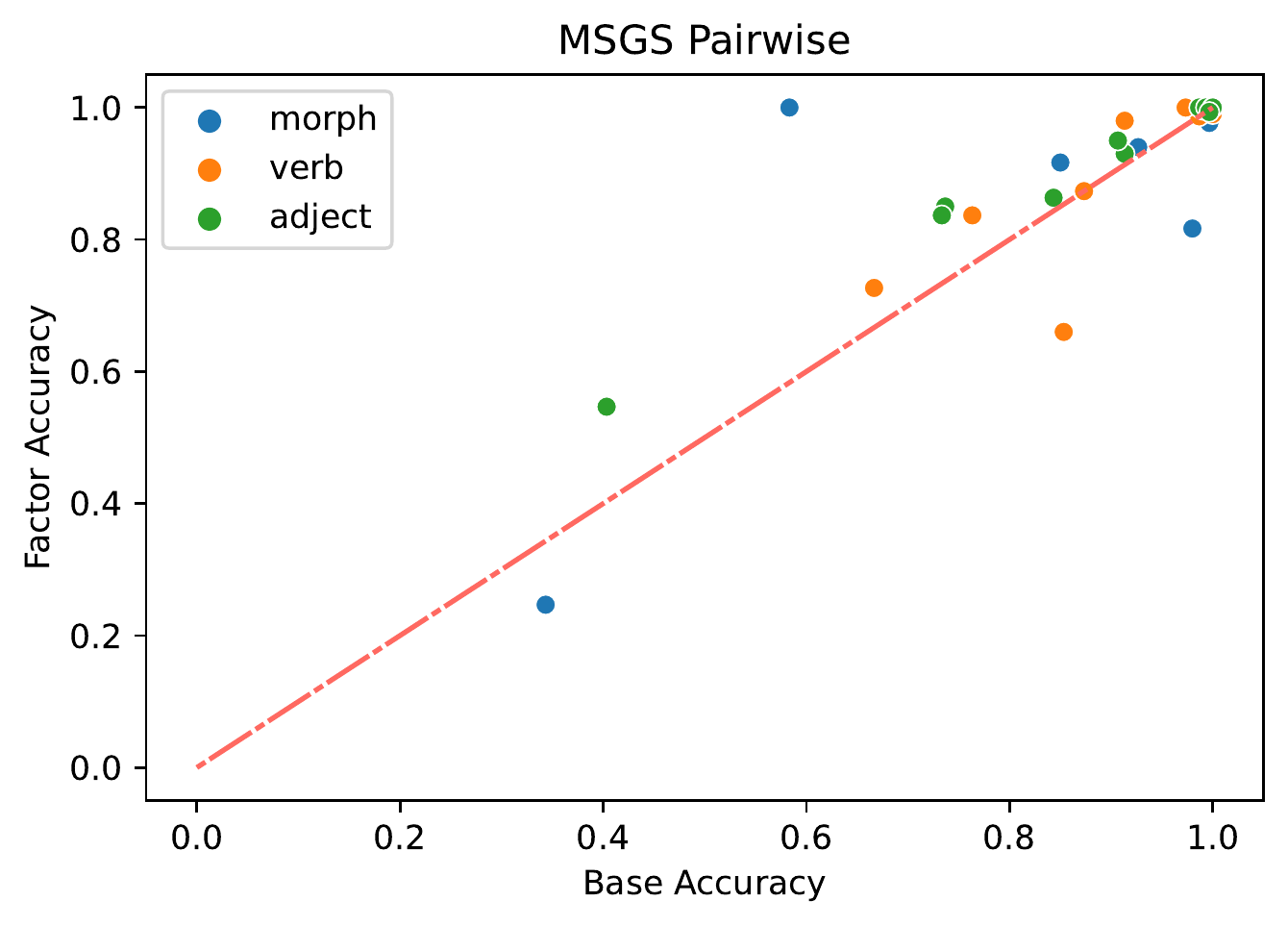}

\caption{LIME pairwise factor against baseline accuracies for \msgs. See Figure~\ref{fig:pairwisecomp} for a related analysis using SHAP.} 
\vspace{-3mm}
\label{fig:msgspairwise}
\end{figure}

\paragraph{Factors differentiate models strongly, though not always in a way aligned with OOD performance.}

Figure~\ref{fig:extremityplot} and Table~\ref{tab:pawspairwiseshap} both show that factors will often consistently decide in favor of a certain model regardless of the choice of $D_{(O, t)}$, especially when dealing with models with different base architectures. Since ranking accuracy correlates with whether these strong alignments are consistent across a spectrum of models and choose the models with higher OOD performance, the tendency for factors to strongly favor a specific model doesn't necessarily correlate with strong overall performance, but does heavily imply that these factors extract meaningful information about the model from the attributions. Looking close at Table~\ref{tab:pawspairwiseshap}, we can see that that even between different model architectures, certain factors are more (\texttt{INDEX-DIFF}) or less (\texttt{SWAP-MAX-DIFF}) capable of making these distinctions.


\paragraph{Factors as projections of model feature space.}

Based on these results, we have evidence that the distributions of attributions are unique to models: in other words, a factor is like a scalar \emph{signature} for a model's feature space with respect to some relevant features. Methods like inoculation, that change a model's behavior in direct ways lead to regular changes in that signature. In these cases, factors align with OOD performance, which explains why factors are so strong in our inoculated experiments. For our non-inoculated experiments (i.g. ELECTRA vs DeBERTa), the feature spaces are fundamentally different, so factor signatures will still capture these differences, but in a way less aligned with ranking on OOD performance. Future work may be able to expand on these differences and what they tell us beyond OOD performance.

\section{Related Work}

This paper relates to a long line of work on understanding explanations, including investigating human ability to interpret explanations~\cite{miller2019, goldberg2020, alqaraawi2020, nguyen2021}, explanations' faithfulness and ability to detect shortcuts~\cite{Geirhos2020} or spurious features~\cite{NLPshortcuts, Madsen2021EvaluatingTF, zhou2021}, and applications to OOD data~\cite{ye-durrett-2022-explanations,choi2022}, including papers in the intersection of multiple directions~\cite{adebayo2022post, kim2021}.


Past work has also investigated using explanations to detect spurious correlations \cite{kim2021,NLPshortcuts,adebayo2022post}. We are different in that we focus on ranking an array of models which exhibits different levels of generalization abilities, as opposed to giving a binary judgment of whether a model is relying on some shortcuts \cite{kim2021,NLPshortcuts,adebayo2022post}. In addition, we experiment with tasks having nuanced shortcuts `in the wild', contrary to synthetically constructed datasets in \citet{NLPshortcuts}. In particular, \citet{adebayo2022post} study the usefulness of explanations in detecting unknown spurious features in an image classification task involving (realistic) possible shortcuts, but find that attributions are ineffective for detecting unknown shortcuts in practice.



\section{Conclusion}

We establish a robust framework for evaluation of fine-grained few-shot prediction of OOD performance, benchmarking approaches in this setting on a range of models. We find that accuracy is a reliable baseline, but intuitive attribution-based factors derived from explanations can sometimes better predict how models will perform in OOD settings, even when they have similar in-domain performance. We further analyze patterns of our approaches, discovering the potential for factors to represent views of model feature space, leaving further exploration to future work.

\section{Limitations}

There are a large number of explanation techniques and many domains these have been applied to. We focus here on a set of textual reasoning tasks like entailment where spurious correlations have been frequently identified. However, correlations in other settings like medical imaging \cite{adebayo2022post} could yield different results. We also note that these datasets are all English-language and use English pre-trained models, so different settings may yield different results; additionally, our factors depend on how explanations are normalized between different examples.

Our paper and analysis themselves comment on the limitations of our methodology as well as explanations as a whole: we find that while explanations often can clearly distinguish different models, knowing which factors will do so, or guaranteeing that explanations align with OOD performance, remains difficult.

\section*{Acknowledgments}

This work was supported by NSF CAREER Award IIS-2145280, a gift from Salesforce, Inc., and a gift from Adobe. The authors acknowledge the Texas Advanced Computing Center (TACC) at The University of Texas at Austin for providing HPC resources used to conduct this research.

\bibliography{anthology,custom}
\bibliographystyle{acl_natbib}

\appendix
\section{Details of Inoculation}
\label{appendix:inoc_details}

One of the methods we used to obtain models with different performances on the OOD sets was inoculation \cite{liu-etal-2019-inoculation}, which involves fine-tuning or further fine-tuning models on small amounts or batches of OOD data alongside in-domain data to bring model performance on OOD sets up. 

\paragraph{MSGS} We borrow notation from~\citet{msgs}.  Most of the fine-tuning data is ambiguous data that doesn't test the spurious correlation, but we add in small percentages of non-ambiguous data where the label favors either the surface or linguistic generalization, tilting the model in that direction. Here, for each set (\verbset{}, \morph{}, \adj{}), we used the following inoculation splits. Linguistic (L) and surface(S) are the features that the inoculation data would favor: 2\% L, 2\% S, mixed (2\% L, 1\% S), (1\% L 2\% S), (2\% L, 2\% S), in addition to no inoculation. The results on $D_O$ are present in Table~\ref{tab:msgsinoc}.

\paragraph{HANS} Specific innoculation results for RoBERTa-large are present in Table~\ref{tab:innoc_details}. We additionally use MNLI pre-trained ELECTRA and DeBERTa models from huggingface. These performance details are also located in Table~\ref{tab:innoc_details}.

\paragraph{PAWS} We used several inoculation techniques to get a variable number of models here. For our RoBERTA-base model, we start with the base model (35\% OOD accuracy) and fine-tune it with $D_T$ data with 2\% of the data having $D_O$ data mixed in. We trained this over several epochs to get models with 82.8\% and 90.8\% accuracy on $D_O$. We also tried fine-tuning our 35\% model on batches of pure $D_O$ data to get a model with 69\% accuracy. For our ELECTRA and DeBERTA models, we use similar batch-only inoculation (fine-tuning on batches of only OOD data). More details are present in Table~\ref{tab:innoc_details}.

\begin{table}[t]
    \centering
    \small
    \begin{tabular}{l|c|c|c}
            \toprule
           &  \multicolumn{1}{c}{\verbset{} \  } & \multicolumn{1}{|c}{\morph{} \ } & \multicolumn{1}{|c}{\adj{} \ } \\
          \midrule
         
         {\sc No-inoc} &  12.0 & 95.0 & 51.0   \\
         {\sc 2L } & 99.0 & 98.0 & 99.2   \\
         {\sc 2S } & 0.0 & 0.0 & 0.0   \\

         {\sc 2L 1S}   & 80.0 & 68.0 & 73.0 \\
         {\sc 1L 2S} & 33.0  & 57.0 & 32.0  \\
         {\sc 2L 2S}  & 53.7 & 49.0 & 56.0   \\
         
         \bottomrule
    \end{tabular}
    \caption{\msgs{} accuracies of various inoculated models. }
    \label{tab:msgsinoc}
\end{table}

 Tables~\ref{tab:hanspawslime}-\ref{tab:hanspawstokig} contain the same information as Table~\ref{tab:hanspawsfewshotshap}, but for the other 2 studied explanation technique.

\begin{table*}[t]
    \centering
    \small
    \begin{tabular}{lccccc}
            \toprule
           Dataset &  \multicolumn{1}{c}{OOD Performance} & \multicolumn{1}{c}{Huggingface Model Name} &
           \multicolumn{1}{c}{LR} &
           \multicolumn{1}{c}{Warmup} &
           \multicolumn{1}{c}{Steps} \\
           
          \midrule
         
         {\sc HANS} &  99.8/99.4 & \texttt{roberta-large-mnli} & $1\text{e}^{-5}$ & 500  & 150   \\
         {\sc HANS } & 96.7/97.6 & \texttt{roberta-large-mnli} & $1\text{e}^{-5}$ & 500 & 100  \\
         {\sc HANS } & 87.1/70.1 & \texttt{roberta-large-mnli} & $1\text{e}^{-5}$  &  500  & 75 \\
         
         {\sc HANS}   & 79.5/62.5 & \texttt{roberta-large-mnli} & $1\text{e}^{-5}$&   500 & 50 \\
         {\sc HANS} & 69.9/58.7  & \texttt{roberta-large-mnli} & $1\text{e}^{-5}$ & $500$ & 25 \\
         {\sc HANS}  & 66.8/57.8 & \texttt{roberta-large-mnli} & $-$ & $-$  & $-$ \\
         {\sc HANS} & 63.5/72.5 & \texttt{howey/electra-base-mnli} & $-$ & $-$ & $-$ \\
         {\sc HANS} & 62.7/65.7 & \texttt{MoritzLaurer/DeBERTa-v3-base-mnli} & $-$ & $-$ & $-$   \\
         \midrule
         
         \msgs{} &  Table~\ref{tab:msgsinoc} & \texttt{roberta-base} & $1\text{e}^{-5}$  & 600 & 6000 \\
         \midrule
         {\sc PAWS} &  90.8 & \texttt{roberta-base} & $1\text{e}^{-5}$ &1200 &12000   \\
         {\sc PAWS } & 82.8 & \texttt{roberta-base} & $1\text{e}^{-5}$ &1200 &12000 \\
         {\sc PAWS } & 69.0 & \texttt{roberta-base} & $1\text{e}^{-5}$  &1200 &7600 \\
         
         {\sc PAWS}   & 35.0 & \texttt{roberta-base} & $1\text{e}^{-5}$&1200 &12000\\
         {\sc PAWS} &  80.5 & \texttt{google/electra-base-discriminator} & $1\text{e}^{-5}$  &1200 &7700 \\
         {\sc PAWS} &  71.8 & \texttt{microsoft/deberta-base} & $1\text{e}^{-5}$  &1200 &7600 \\
         \bottomrule
    \end{tabular}
    \caption{Architecture details for our experiments. ``Steps'' indicates the number of gradient updates from the specified dataset that are applied to the model. For HANS models, performance is on \hanssub/\hanscon. For all models, small batch sizes were used, with weight decay of $0.1$.}\label{tab:innoc_details}
\end{table*}

\begin{table}[t]
    \centering
    \small
    \begin{tabular}{l|c|c|c}
            \toprule
           &  \multicolumn{1}{c}{\paws{}} & \multicolumn{1}{c}{\hanssub} & \multicolumn{1}{c|}{\hanscon{}} \\
          \midrule
         \bf Baselines    \\
         {\texttt{ACCURACY}} &  88.7 & 90.6 & 81.6   \\
         {\texttt{CONFIDENCE} } &  9.2 & 40.4 & 52.8   \\
         {\texttt{RANDOM} } & 50.7 & 51.4 & 49.6   \\
         
         \midrule
         \bf Explanations   \\
        
         {\texttt{CONST}} & $-$  & $-$ & 79.4  \\
         {\texttt{SWAP-MAX-DIFF}}  & 80.6 & $-$ & $-$   \\
         {\texttt{SWAP-AVG}} & 98.2 & $-$ & $-$  \\
         \midrule
         {\texttt{MAX-DIFF}} & 70.1 & 67.2 & 58.3   \\
         {\texttt{INDEX-DIFF}} & 70.5  & 88.5 & 67.2   \\
         {\texttt{SUM-DIFF}} &  53.9 & 59.1 & 60.4   \\
         {\texttt{FIRST-TOK}} & 55.4 & 51.0  & 81.3   \\
         \bottomrule
    \end{tabular}
    \caption{LIME version of Table~\ref{tab:hanspawsshapother}}
    \label{tab:hanspawslime}
\end{table}

\begin{table}[t]
    \centering
    \small
    \begin{tabular}{l|c|c|c}
            \toprule
           &  \multicolumn{1}{c}{\paws{}} & \multicolumn{1}{c}{\hanssub} & \multicolumn{1}{c|}{\hanscon{}} \\
          \midrule
         \bf Baselines    \\
         {\texttt{ACCURACY}} &  88.7 & 90.6 & 81.6   \\
         {\texttt{CONFIDENCE} } & 9.2 & 40.4 & 52.8   \\
         {\texttt{RANDOM} } & 50.7 & 51.4 & 49.6   \\
         
         \midrule
         \bf Explanations   \\
        
         {\texttt{CONST}} & $-$  & $-$ & 79.2  \\
         {\texttt{SWAP-MAX-DIFF}}  & 84.3 & $-$ & $-$   \\
         {\texttt{SWAP-AVG}} & 85.6 & $-$ & $-$  \\
         \midrule
         {\texttt{MAX-DIFF}} & 86.9 & 55.2 & 69.9   \\
         {\texttt{INDEX-DIFF}} & 51.4 & 85.8 & 53.4   \\
         {\texttt{SUM-DIFF}} &  51.6 & 77.4 & 68.1   \\
         {\texttt{FIRST-TOK}} & 64.0 & 69.7  & 59.2   \\
         \bottomrule
    \end{tabular}
    \caption{Tokig numbers for Table \ref{tab:hanspawsshapother}}
    \label{tab:hanspawstokig}
\end{table}

\section{Bootstrapping Details}
\label{appendix:bootstrapdets}

We now describe our process for bootstrapping and evaluating the capability of explanations in our setting.

For a sampled population of examples from the $D_O$ set, for the $m$ models that we're examining at a time, we generate explanations for each of the $m$ models on all of the sampled population. We then repeatedly take a sample with replacement (500 times) of 10 examples $D_{O, t}$ each, where we have $500 \times 10 \times m$ total explanations we want to examine. We calculate factors for each of the 10 explanations in each $D_{O, t}$ sample and pool them to get a list of factor metrics for the $D_{O, t}$, one for each model. 

For each pair, we then look at the ground-truth $D_O$ ranking for models and their respective factor metrics, getting \emph{successes} where these match, and \emph{failures} otherwise. When we average these accuracies across our 500 bootstrap samples, we get pairwise distributions (the distribution of successes vs failures on a sample for a given pair), which we can further aggregate to get few-shot accuracies. 

Note, in practice, to prevent variance from run-to-run, we fix the population of 500 $D_{O, t}$s, but we validated that re-running on new sampled populations didn't impact any numbers greatly. Though we tried using several (5, 10, 20) $D_{O, t}$ sizes, we decided to use the probe size of 10 as a realistic probe size for our setting, which wouldn't be burdensome to hand-craft in practice.

Our methodology can be run quickly in a post-hoc manner as many times as needed on top of a population of the necessary explanations. 

\begin{table}[t]
    \centering
    \small
    \begin{tabular}{clcc}
            \toprule
         \multicolumn{2}{l}{ID Set \hspace{1.5em} OOD Set}  &  {$D_O$ Size} & {$D_{(O, t)}$ Size}\\
         \midrule
         \multirow{3}{*}{\sc{msgs}}
         & \morph{}   & 10000 & 10 \\
         & \verbset{} & 10000 & 10\\
         & \adj{} & 10000 & 10\\
         \midrule
         \multirow{2}{*}{\sc{mnli}}
         & \hanssub{}  & 10000 & 10 \\
         & \hanscon{} & 10000 & 10\\
         \midrule
         \multirow{1}{*}{\sc{ qqp}}
         & \paws{}  & 677 & 10\\
         \bottomrule
    \end{tabular}
    \caption{Information regarding our considered datasets. For all datasets, the bootstrap sample size is fixed at 10.}\label{tab:data_settings}
\end{table}

\section {Additional Plots}

Figure~\ref{fig:extremityplot} shows additional information about the distrbution of pairwise accuracies between different model architectures.

\begin{figure}[t!]
\centering
\includegraphics[width=0.48\textwidth]{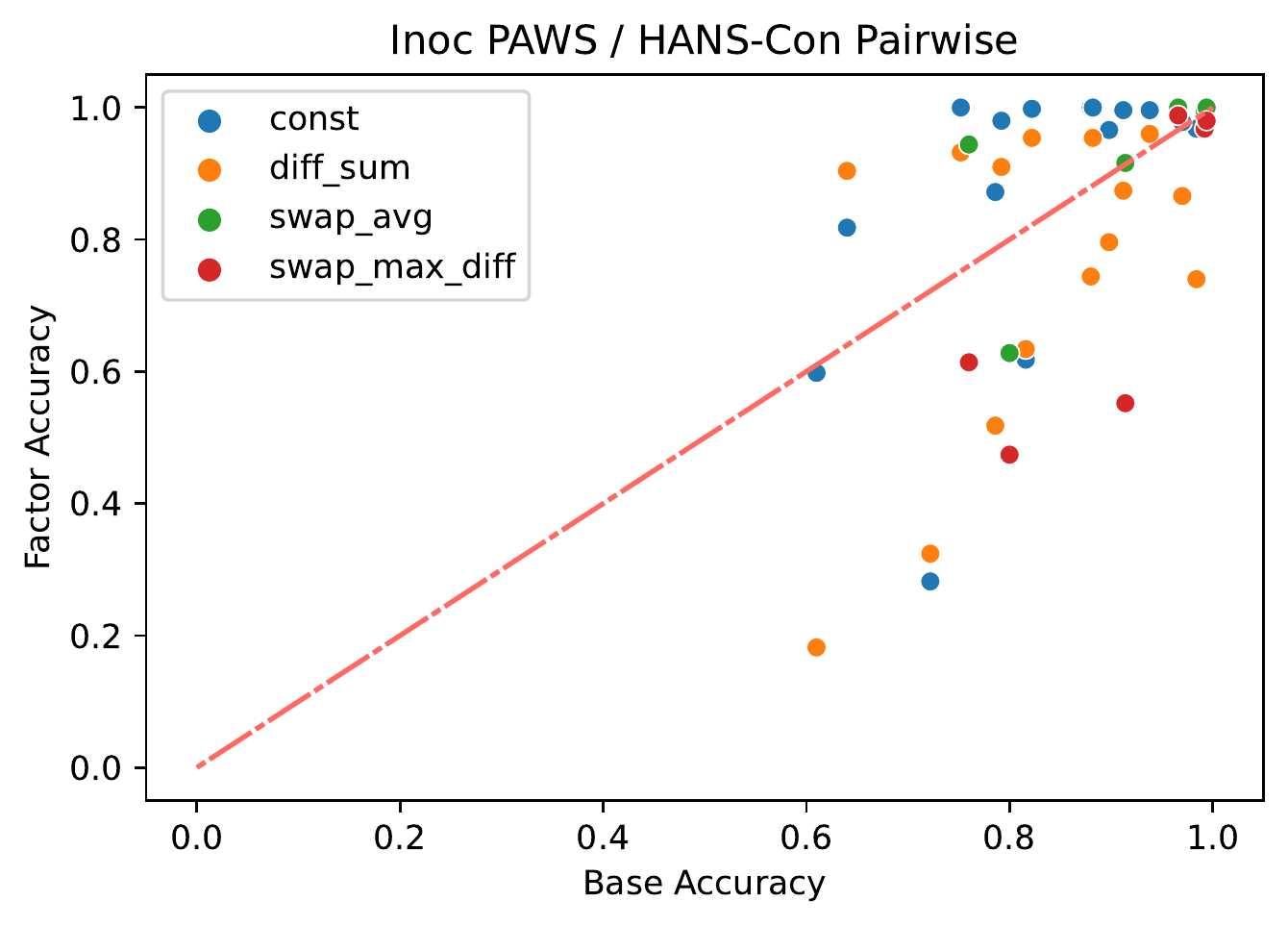}
\caption{SHAP pairwise factor compared to \accuracy{} for \hanscon{} and \paws{}. Each point represents a factor accuracy (y-axis) for a pair of models in comparison to \accuracy{} (x-axis) for the same pair. Points above the red $y=x$ line represent factors outperforming the accuracy baseline. \texttt{CONST} and \texttt{DIFF-SUM} are for \hanscon{}, \texttt{SWAP-AVG} and \texttt{SWAP-MAX-DIFF} are for \paws{}} 
\vspace{-3mm}
\label{fig:pairwisecomp}
\end{figure}

\section{Reproducibility}
\subsection{Computing Infrastructure}
All experiments were conducted on a desktop with 2 NVIDIA 1080 Ti (11 GB) and 1 NVIDIA Titan Xp (12 GB).

\subsection{Runtimes}
For PAWS and MSGS fine-tuned models, we fine-tuned for roughly 1 GPU hour per model. Since HANS models were trained for very few steps, their training time is inconsequential. Generating attributions required for numerical evaluation took less than 6 GPU hours.


\subsection{Dataset Details}
We used datasets in the JSONL format. We simplified all our dataset settings to binary classification for simplicity, and used data directly from the downloads made available in the original papers.

\end{document}